\documentclass{article} 
\usepackage[table]{xcolor}
\usepackage{_sty/iclr2026_conference,times}

\usepackage{hyperref}
\usepackage{url}

\usepackage{booktabs}
\usepackage{graphicx}
\usepackage{multirow}
\usepackage{diagbox}
\usepackage{makecell}
\usepackage{pifont}
\usepackage{amsfonts}
\usepackage{subcaption}
\usepackage{wrapfig}
\usepackage{amsmath}
\usepackage{extarrows}
\usepackage{caption}
\usepackage{float}
\usepackage{algorithm}
\usepackage{algorithmic}
\usepackage{cleveref}
\usepackage{color}

\definecolor{mygray}{RGB}{230, 230, 230}

\title{ToProVAR: Efficient Visual Autoregressive Modeling via Tri-Dimensional Entropy-Aware Semantic Analysis and Sparsity Optimization}

\author{Jiayu Chen$^{1}$
\quad
Ruoyu Lin$^{2}$
\quad
Zihao Zheng$^{1}$
\quad
\textbf{Jingxin Li$^{2}$} \\
\textbf{Maoliang Li$^{1}$}
\quad
\textbf{Guojie Luo$^{1,3}$}
\quad
\textbf{Xiang Chen$^{1}$}\thanks{Corresponding author.} \\
$^{1}$ School of Computer Science, Peking University \\
$^{2}$ School of Electronics Engineering and Computer Science, Peking University \\
$^{3}$ National Key Laboratory for Multimedia Information Processing, Peking University
}

\newcommand{\XC}[1]{\ifbool{inccomment}{{\color{magenta}YC\@: #1}}{}}
\newcommand{\JY}[1]{\ifbool{inccomment}{{\color{blue}XC\@: #1}}{}}
\newcommand{\TD}[1]{\ifbool{inccomment}{{\color{orange}#1}}{}}
\newcommand{\FN}[1]{\ifbool{inccomment}{{\color{OliveGreen}#1}}{}}

\iclrfinalcopy 
\begin{document}

\maketitle

\vspace{-5mm}
\begin{abstract}
    Visual Autoregressive (VAR) models enhance generation quality but face a critical efficiency bottleneck in later stages. 
    In this paper, we present a novel optimization framework for VAR models that fundamentally differs from prior approaches such as FastVAR and SkipVAR. Instead of relying on heuristic skipping strategies, our method leverages attention entropy to characterize the semantic projections across different dimensions of the model architecture. This enables precise identification of parameter dynamics under varying token granularity levels, semantic scopes, and generation scales. Building on this analysis, we further uncover sparsity patterns along three critical dimensions—token, layer, and scale—and propose a set of fine-grained optimization strategies tailored to these patterns. Extensive evaluation demonstrates that our approach achieves aggressive acceleration of the generation process while significantly preserving semantic fidelity and fine details, outperforming traditional methods in both efficiency and quality.
    Experiments on Infinity-2B and Infinity-8B models demonstrate that ToProVAR achieves up to 3.4× acceleration with minimal quality loss, effectively mitigating the issues found in prior work. Our code will be made publicly available.

\end{abstract}

\begin{figure}[H]
    \includegraphics[width=\linewidth]{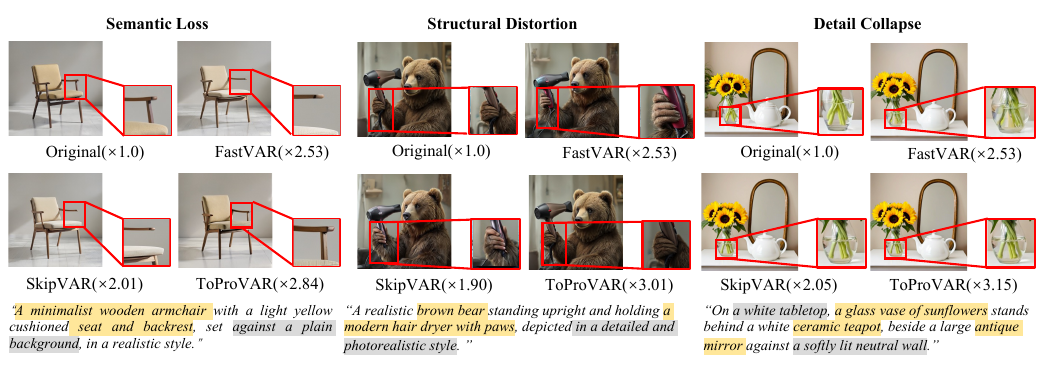}
    \caption{A comparison between our method and state-of-the-art compression methods. The SOTA methods often suffer from issues such as semantic loss, structure distortion, and detail collapse.}
    \label{fig:1}
\end{figure}

\section{Introduction}
\label{sec:intr}

Traditional autoregressive (AR) models generate images via raster-scan next-token prediction~\citep{li2024mar, liu2024lumina, SandAI2025MAGI1, xie2024showo}, which has long produced inferior results compared to diffusion models. Visual AutoRegressive modeling (VAR)~\citep{tian2024var,han2024infinity, tang2024hart} reformulates generation as coarse-to-fine next-resolution prediction, enabling GPT-style AR models to, for the first time, surpass diffusion models in image quality. Despite these advancements, VAR-based methods still face a core problem: the number of tokens grows exponentially with image resolution and generation scales, resulting in inefficient computation in later stages.

\begin{figure*}[t]
    \vspace{-6pt}
    \includegraphics[width=\linewidth]{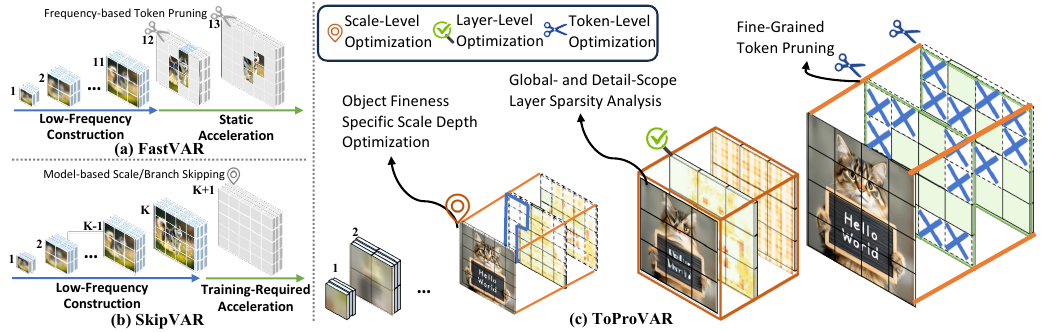}
    \caption{Different Optimization Dimensions -- FastVAR vs. SkipVAR vs. ToProVAR}
    \label{fig:2}
    \vspace{-2mm}
\end{figure*}

To improve the computational efficiency of VAR models, existing researchs, such as FastVAR~\citep{guo2025fastvar} and SkipVAR~\citep{li2025skipvar}, have explored various token reduction strategies. As shown in Fig.~\ref{fig:2}(a)(b), FastVAR retains a fixed ratio of high-frequency tokens in the token dimension, while SkipVAR skips certain scales or replaces unconditional branches in the scale dimension based on a trained decision model. Although these methods have demonstrated significant value in accelerating generation, they mainly rely on single-dimensional sparsity analysis of intermediate image data, which introduces several limitations as illustrated in Fig.~\ref{fig:1}: (1)Semantic loss: specific tokens corresponding to key objects are pruned when their frequency in the image is too low; (2) Structural distortion: a single sparsity metric like frequency in FastVAR cannot capture the complex relative relationships among objects, leading to noticeable deformations in complex regions; (3) Detail collapse: fine-grained objects typically require deeper generation scales for adequate support, while methods like SkipVAR that skip entire scales result in severe loss of details.

Based on the above issues, we identify several critical challenges in optimizing VAR models: (1) Fine-grained sparsity analysis: Unlike prior work, we need to design a highly fine-grained approach to sparsity analysis that effectively prevents information loss caused by misalignment between sparsity metrics and image semantics. (2) Multi-dimensional representation: It is essential to analyze the model across multiple dimensions, enabling not only the assessment of individual token importance but also the accurate characterization of their relative relationships in other dimensions. (3) Efficiency-preserving optimization: While pursuing fine-grained optimization, the analysis itself must remain efficient; otherwise, excessive overhead in modeling sparsity would compromise the overall benefits of the optimization.

To address the aforementioned challenges, we introduce a novel optimization framework -- \textbf{ToProVAR} -- for VAR modeling computing optimization.

\hspace{6mm} First, unlike prior approaches such as FastVAR or SkipVAR, which evaluate sparsity directly on intermediate image representations, we leverage attention entropy to analyze how semantics are projected within the model structure during generation. This enables precise tracking of dynamics under varying object salience, semantic scope, and fineness, thereby providing principled guidance for joint semantic–sparsity analysis and optimization.

\hspace{6mm} Second, we extend entropy-based analysis beyond tokens to cover three complementary dimensions: token-level, layer-level, and scale-level. This multi-dimensional perspective allows us to uncover semantic distributions and correlations that govern sparsity patterns along token, layer, and scale dimensions. As illustrated in Fig.~\ref{fig:2}(c), this facilitates a series of fine-grained optimizations: token-level pruning of non-essential semantics, layer-level compression that distinguishes global from detail representation, and scale-level disentanglement and depth adjustment of generation tailored to object fineness necessity.

\hspace{6mm} Finally, we design a coordinated optimization algorithm that integrates these three dimensions, while further improving the efficiency of attention-entropy analysis itself. This ensures not only the effectiveness but also the practicality of our framework for real-world VAR acceleration.

We evaluated ToProVAR on mainstream VAR models, Infinity-2B and Infinity-8B. The experimental results show that, compared to FastVAR and SkipVAR, ToProVAR improves inference speed to nearly 3.5× with almost no loss in image quality. In Fig.~\ref{fig:1}, we demonstrate the high-quality visual results generated by ToProVAR based on the Infinity-8B model, effectively addressing issues such as semantic loss, structural distortion, and detail collapse.

\section{Preliminary}
\label{sec:preliminary}

\textbf{Visual Autoregressive Modeling.}
VAR redefines the traditional autoregressive paradigm, shifting the core from ``next-token prediction'' to ``next-scale prediction.''
    For a given image, VAR first obtains a feature map through an encoder, which is then quantized into $K$ multi-scale token maps $\mathcal{R} = \{r_1, r_2, \dots, r_K\}$; 
    The resolutions of these token maps increase progressively according to a scale schedule $\{(h_1, w_1), (h_2, w_2), \dots, (h_K, w_K)\}$. 
    Each token map $r_k \in \{1, \dots, V\}^{h_k \times w_k}$ contains $h_k \times w_k$ discrete tokens, all from a codebook of size $V$.

The joint probability distribution over the multi-scale token maps is factorized autoregressively as:
\begin{equation}
    p(r_1, r_2, \dots, r_K) = \prod_{k=1}^{K} p(r_k | r_1, r_2, \dots, r_{k-1}).
    \label{eq:var_factorization}
\end{equation}
    Specifically, at each autoregressive scale $k$, the previously generated token maps $\{r_1, \dots, r_{k-1}\}$ are processed by $L$ layers of the VAR network to predict the current token map $r_k$. 
    This multi-scale generation process supports parallel decoding of multiple tokens within a single token map, significantly improving efficiency compared to traditional ``token-by-token'' autoregressive models.

\textbf{Attention Entropy for Semantic Projection Analysis.}
Attention entropy quantifies how concentrated the attention distribution is for a given query. A low entropy value indicates that a token focuses its attention on only a few targets, suggesting strong semantic selectivity; conversely, high entropy reflects a more uniform distribution over targets, implying weaker semantic focus. Formally, given a query $q_i$ and keys ${k_j}$, the attention weights are defined as scaled dot-products, and the entropy is computed as:
\begin{equation}
\mathcal{H}(q_i) = -\sum_{j=1}^{N} \alpha_{i,j} \log \alpha_{i,j},
\quad
\alpha_{i,j} = \frac{\exp(q_i^\top k_j / \sqrt{d_k})}{\sum_{l=1}^{N} \exp(q_i^\top k_l / \sqrt{d_k})},
\label{eq:attn_entropy_def}
\end{equation}
where $q_i$ and $k_j$ denote the query and key vectors, and $d_k$ is their dimension.
Previous studies \citep{cheng2022entropy, pardyl2023active, choi2024vid} have exploited this property of attention entropy to distinguish between foreground and background regions. In this work, we build upon this intuition and employ attention entropy for a different purpose: performing fine-grained semantic projection evaluation of VAR models.

Unlike frequency-based fixed scopes, however, the range of $j$ can be flexibly set across different model structures. 
    Extending $j$ beyond local tokens to encompass cross-layer and multi-scale representations enables entropy to jointly capture semantic evolution across layers and scales.
    This allows us not only to analyze variations between adjacent tokens, but also to extend the activation range across layer and scale dimensions. 
In this way, we can simultaneously preserve fine-grained token-level analysis while examining sparsity from broader perspectives in the generation process.

\section{Tri-Dimensional Attention Entropy Generalization and Related Semantic and Sparsity Analysis}
\label{sec:anal}

\begin{figure*}[t]
    \includegraphics[width=\linewidth]{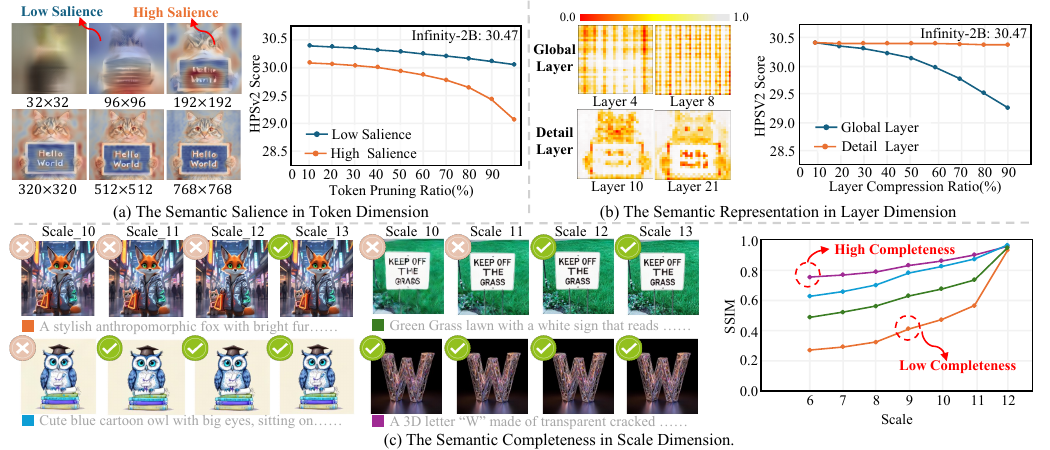}
    \caption{
Tri-dimensional attention entropy analysis in VAR models:  
(a) \textbf{Token-Level Semantic Salience}: Pruning low-saliency tokens preserves quality, while pruning high-saliency ones causes severe degradation.  
(b) \textbf{Layer-level Semantic Representation}: Global Layers capture structure and are pruning-sensitive, whereas \emph{detail layers} refine local semantics and can be pruned.  
(c) \textbf{Scale-level Semantic Depth}: complex objects require deeper scales for fine details, while simple ones stabilize earlier, enabling adaptive depth pruning.
}
    \label{fig:3}
\end{figure*}

We leverage attention entropy as a unified measure for semantic projection evaluation, enabling the analysis of data sparsity from three dimensions in VAR models.

\textbf{Token-Level Attention Entropy for Semantic Salience Analysis.}
Previous approaches, such as frequency-based FastVAR, often overlook semantic information; as a result, tiny and fine details are easily over-optimized and consequently lost, as illustrated in Fig.~\ref{fig:1}. In contrast, by employing attention entropy, our analysis is performed at the model dimension, which enables us to effectively capture generation semantics even for fine-grained content. This substantially improves both controllability and accuracy in generation optimization. Frequency-based approaches typically rely on local averaging within a region. When most of the region contains low-frequency content but only a small portion carries high-frequency details, the averaging process suppresses the latter, leading to their removal during pruning (e.g., the cat’s paw in the figure). By leveraging semantic cues, our semantic-based method overcomes this limitation and faithfully preserves both local details and global structures, even under high token pruning ratios. 

Therefore, semantic salience refers to the salience distribution across tokens, where only a subset of tokens carries critical semantic information.
To further evaluate the impact of semantic salience on quality, we progressively pruned tokens from regions of different salience and measured image quality.
    The results, shown in the right of Fig.~\ref{fig:3}(a), demonstrate that on the Infinity-8B model, removing up to 90\% of low-saliency tokens leads to only a slight decline in image quality (a drop of $<$2\%).
    However, pruning just 10\% of high-saliency tokens causes a significant loss in generation quality, resulting in noticeable artifacts and semantic gaps.
    
Based on this exploration of semantic salience, we propose prioritizing the pruning of low-saliency token regions during the later stages of generation. This approach can accelerate the generation process while preserving the quality of the output.

\textbf{Layer-level Attention Entropy for Semantic Scope Analysis.}
Attention entropy not only enhances granularity at the token level, but its flexible analysis scope also enables semantic patterns to be examined across broader dimensions. By extending the scope of the attention entropy (the range of index $j$ in Eq.~\ref{eq:attn_entropy_def}) to include all tokens within a layer, we can analyze token distributions in the layer dimension and characterize the generation scope of each layer. Specifically, as shown in Fig.~\ref{fig:3}(b), certain layers exhibit relatively uniform, grid-like attention distributions with prominent principal components, focusing on capturing broad spatial relationships and establishing the overall image structure at a global scope. In contrast, the other layers display more varied, semantic-driven attention patterns with less distinct principal components, concentrating on progressively refining local semantics and fine-grained textures within a localized scope. 

This observation motivates distinguishing layers according to their semantic scope, categorizing them as Global Layers and Detail Layers. \emph{Global layers}, with their globally distributed attention, encode strong interconnections across the entire image, whereas \emph{detail layers}, with more localized attention, concentrate on specific regions, leaving substantial sparsity in the unattended areas. Fig.~\ref{fig:3}(b) further illustrates the impact of semantic scope on generation quality: compressing \emph{global layers} by more than 50\% significantly degrades output quality, whereas even aggressive compression of \emph{detail layers}—up to 90\% on the Infinity-8B model—maintains high fidelity. This indicates that selectively pruning \emph{detail layers} can accelerate generation with minimal quality loss.

Implementing this strategy, however, requires a reliable method to distinguish global from detail layers. To this end, we propose to identify global and detail layers by the prominence of principal components in their attention entropy distributions and then prune only the \emph{detail layers}.

\textbf{Scale-level Attention Entropy for Semantic Fineness Analysis.}
VAR models generate images across multiple autoregressive (AR) scales. Prior approaches often applied coarse-grained scale reduction to reduce computation, but this came at the cost of fine-grained quality. As shown in Fig.~\ref{fig:1}, images containing fine details require optimization at the semantic granularity level rather than purely at the scale level.

Extending the scope of attention entropy (Eq.~\ref{eq:attn_entropy_def}) beyond local tokens to multiple scales enables us to capture the semantic evolution across scales. Specifically, images with high fineness, such as a complex``cyber fox'', exhibit predominantly low-salience distributions, requiring deeper scales to render fine details. In contrast, simpler objects like the letter ``W'' show predominantly high-salience distributions, stabilizing at shallower scales. As shown in Fig.~\ref{fig:3}(c), this trend is further corroborated by their SSIM curves.

This scale-level evaluation of attention entropy thus provides a principled way to distinguish the fineness requirements of different objects and adapt the depth accordingly. Based on this observation, we propose dynamically determining the starting scale for pruning according to semantic fineness, enabling adaptive depth allocation for different generation tasks.

\section{Tri-Dimensional VAR Optimization}
\label{sec:method}

\begin{figure*}[t]
    \includegraphics[width=\linewidth]{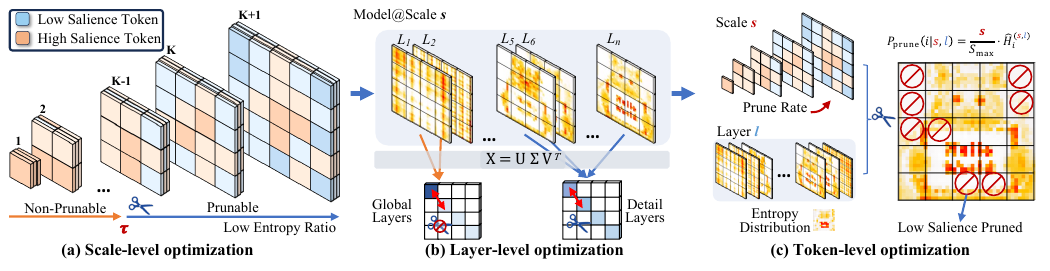}
    \caption{
Tri-Dimensional Entropy-Aware VAR Sparsity Optimization:
(a) \textbf{Scale-level}: compute the low-entropy ratio $\rho_s$ across scales and select the pruning start depth via threshold $\tau$.
(b) \textbf{Layer-level}: for each scale, perform SVD on the entropy map to separate \emph{Global} layers from \emph{Detail} layers.
(c) \textbf{Token-level}: within prunable layers/scales, increase pruning rate with scale and use entropy-based gating $p_{\text{prune}}$ to remove low-salience tokens, preserving salient regions.
}
    \label{fig:4}
\end{figure*}

Based on our tri-dimensional generalization of attention entropy and the corresponding semantic analysis, we propose a comprehensive framework with optimization techniques for each dimension. 
    As illustrated in Fig.~\ref{fig:4}, given the multi-scale token maps of a VAR model, $\mathcal{R} = \{r_1, r_2, \dots, r_k\}$, the framework optimizes the generation process across three dimensions. 
    Specifically, it first estimates the semantic fineness of the scales $\mathcal{R}$ to dynamically determine the scale depth for pruning, $r_i$. 
    Subsequently, within applicable scales, it analyzes each layer's semantic scope to distinguish between Global and Detail layers, pruning only the latter. 
    Finally, fine-grained token-level sparsification is performed within these layers based on attention entropy, with a unified gating function $G$ integrating information from all dimensions to determine the pruning probability for each token.

\textbf{Scale-level optimization via Semantic Fineness.}
Drawing from our scale-level analysis, as the scale increases, the number of high-salience tokens (orange) gradually decreases, implying fewer tokens need to be processed while the semantic fineness of the generated image improves. To capture this effect, we quantify the distribution of high-salience tokens across scales by the proportion of low-entropy tokens: a higher proportion indicates finer semantics.
As shown in Fig.~\ref{fig:4}(a), we define the low-entropy ratio at scale $s$ as
\begin{equation}
\rho_s = \frac{\left| { i \mid H_i^s < \overline{H}^s } \right|}{N_s},
\label{eq:low_entropy_ratio_mean}
\end{equation}
where $H_i^s$ denotes the attention entropy of token $i$ at scale $s$, $\overline{H}^s$ denotes the mean entropy at scale $s$, and $N_s$ denotes the total number of tokens.

Based on this measure, we determine the pruning start scale using a threshold $\tau$. Specifically, the scale depth discrimination function is defined as
$D = \min { s \mid \rho_s \ge \tau }$,
where $D$ is the minimum scale index at which semantic stability is achieved.
To calibrate $\tau$, we conduct multiple pre-sampling experiments. We observe that as generation converges, $\rho_s$ stabilizes, which provides a reliable criterion for dynamically selecting the pruning depth. 

\textbf{Layer-level optimization via Semantic Scope.}
Based on our layer-level analysis, the next optimization step is to distinguish between \emph{Global Layers} and \emph{Detail Layers} according to their semantic scope. As shown in Fig.~\ref{fig:4}(b), Global Layers exhibit a pronounced dominant component, while \emph{Detail Layers} do not. To quantify this, we apply singular value decomposition (SVD) to the attention entropy map $X$ of each layer, i.e., $X = U \Sigma V^{\top}$. In Global Layers, the gap between the largest and the second largest singular values in $\Sigma$ is pronounced, whereas Detail Layers lack such dominance. We thus define the \textit{principal component ratio} $\varrho^{(l,s)} = \sigma^{(l,s)}_1 / \sigma^{(l,s)}_2$, where $\sigma^{(l,s)}_1$ and $\sigma^{(l,s)}_2$ denote the two largest singular values of the token representation matrix at layer $l$ and scale $s$.

If $\varrho^{(l,s)} \gg 1$, the layer is classified as Global; if $\varrho^{(l,s)} \approx 1$, it is considered Detail. To provide a continuous score for pruning decisions, we further define the \textit{layer representation score}:
\begin{equation}
\mathcal{R}^{(l,s)} = \exp\big(-\beta (\varrho^{(l,s)} - 1)\big), \quad \beta > 0,
\end{equation}
where $\mathcal{R}^{(l,s)} \to 1$ for Detail Layers and $\mathcal{R}^{(l,s)} \to 0$ for Global Layers. This score offers a quantitative criterion for layer-level pruning.

\textbf{Token-level optimization via Fine-grained Semantic Salience.}
After identifying the prunable scales and layers, the final stage is to perform fine-grained token pruning based on semantic salience. 
The core idea is to eliminate tokens with low salience (i.e., high attention entropy). 
As illustrated in Fig.~\ref{fig:4}(c), to establish a consistent pruning basis across different layers and scales, we first normalize the token's attention entropy:
\begin{equation}
\hat{H}_{i}^{(l,s)} = \frac{H_{i}^{(l,s)}}{\sum_{j=1}^{N_{s,l}} H_{j}^{(l,s)}}, 
\qquad \text{where} \quad \sum_{i=1}^{N_{s,l}} \hat{H}_{i}^{(l,s)} = 1.
\end{equation}

We then integrate this normalized entropy with the layer score $\mathcal{R}^{(l,s)}$ and a monotonic scale factor $\phi(s) = s/S_{\max}$ to define a unified pruning tendency:
$q_i^{(s,l)} = \phi(s)\cdot \mathcal{R}^{(l,s)} \cdot \hat{H}_i^{(s,l)}$.
This formulation smoothly incorporates all three dimensions, ensuring that tokens with higher entropy, in layers with broader semantic scope, and at deeper scales are more likely to be pruned. 

Finally, the pruning tendency is mapped to a retention probability:
\begin{equation}
P_{\text{keep}}(i \mid s,l) = 
\begin{cases} 
    1, & \text{if } s < D, \\[6pt] 
    1 - \operatorname{clip}\!\Big( \alpha_{\min} + (\alpha_{\max}-\alpha_{\min})\, q_i^{(s,l)},\; 0,\; 1 \Big), & \text{otherwise.}
\end{cases}
\end{equation}

This three-factor integration provides a coherent pruning policy across tokens, layers, and scales, effectively discarding redundant details while preserving semantically critical structures.

\textbf{Computational Optimization for Attention Entropy.}
The attention entropy is formally defined in Equation~\ref{eq:attn_entropy_def}. 
A straightforward implementation would require explicitly materializing the full $N \times N$ attention matrix in order to compute row-wise probability distributions and their entropy. 
However, such an approach is computationally prohibitive in practice, since modern attention implementations such as FlashAttention never instantiate the dense matrix explicitly due to  memory and runtime constraints.

To address this challenge, we extend the original FlashAttention algorithm with an online entropy computation mechanism, which we refer to as \textit{Flash Attention Entropy}. 
The key idea is to preserve the memory efficiency of FlashAttention while simultaneously maintaining sufficient statistics for entropy computation. 
Inspired by the online softmax strategy in FlashAttention, we design an incremental update scheme that avoids forming the full attention matrix.
More concretely, recall that the entropy involves terms of the form $x \log x$ over normalized attention scores. 
By leveraging the algebraic identity
$kx \log(kx) = kx \log x + (\log k) \cdot xk$
We can decompose the entropy computation into two accumulative statistics: 
the standard normalization terms (rowmax $m$ and expsum $l$) that are already tracked in FlashAttention, 
and an additional intermediate statistic that maintains $x \log x$. 
This decomposition ensures that the entropy can be computed in a streaming fashion with negligible overhead relative to the baseline FlashAttention kernel.
The resulting algorithm, termed \textit{Flash Attention Entropy}, thus inherits the linear-time and memory-efficient properties of FlashAttention while enabling exact entropy computation without approximation.

\newcommand{\tbf}[1]{\textbf}
\section{Experiments}

\subsection{Experimental Setup}

\begin{table}[t]
\vspace{-6pt}
\caption{\textbf{Quantitative comparison} on GenEval and DPG. 
Note, GenEval follows the official protocol without rewritten prompts. 
Latency is measured on a single GPU with batch size~1.}
\label{tab:geneval_dpg}
\centering
\small
\setlength{\tabcolsep}{1.5pt}
\begin{tabular}{lcccccccccc}
\toprule
\multirow{2}{*}{Methods} & \multicolumn{4}{c}{GenEval} & \multicolumn{4}{c}{DPG} & \multirow{2}{*}{Latency(s)$\downarrow$} & \multirow{2}{*}{Speedup} \\
\cmidrule(lr){2-5} \cmidrule(lr){6-9}
& Two Obj. & Position & Color Attri. & Overall$\uparrow$ & Entity & Relation & Attribute & Overall$\uparrow$ &  &  \\
\midrule
Infinity-2B & 79.01 & 24.00 & 58.00 & 0.69 & 90.81 & 88.19 & 87.89 & 83.41 & 2.10 & 1.0 $\times$ \\
+FastVAR    & 78.79 & 27.75 & 59.50 & 0.68 & 88.86 & \textbf{91.57} & 87.46 & \textbf{83.39} & 0.80 & 2.6 $\times$ \\
+SkipVAR    & 76.77 & 26.50 & 57.50 & 0.67 & \textbf{89.30} & 87.07 & 87.01 & 82.94 & 1.10 & 2.0 $\times$ \\
\rowcolor{gray!15}
+ToProVAR   & \textbf{78.80} & \textbf{29.50} & \textbf{62.00} &  \textbf{0.69} & 87.39 & 88.92 & \textbf{90.01} & 83.07 & \textbf{0.61} & \textbf{3.4 $\times$} \\
\midrule
Infinity-8B & 96.97 & 61.00 & 75.00 & 0.83 & 90.92 & 93.57 & 88.83 & 86.68 & 4.86 & 1.0 $\times$ \\
+FastVAR    & 94.19 & 57.00 & 75.25 & 0.81 & 90.80 & \textbf{92.30} & 90.40 & 86.50 & 2.01 & 2.4 $\times$ \\
+SkipVAR    & 94.94 & 57.50 & \textbf{76.50} & 0.82 & 89.71 & 90.52 & 90.02 & 86.44 & 2.11 & 2.3 $\times$ \\
\rowcolor{gray!15}
+ToProVAR   & \textbf{94.95} & \textbf{61.00} & 76.00 & \textbf{0.83} & \textbf{91.11} & 90.39 & \textbf{91.04} & \textbf{86.70} & \textbf{1.78} & \textbf{2.7 $\times$} \\
\bottomrule
\end{tabular}
\end{table}

\begin{table}[t]
\caption{\textbf{Quantitative comparison} on HPSv2.1 and ImageReward, two human preference benchmarks. Latency is measured on a single GPU with batch size $1$.}
\label{tab:hpsv2_imagereward}
\centering
\small
\setlength{\tabcolsep}{4pt}
\begin{tabular}{lccccccccc}
\toprule
\multirow{2}{*}{Methods} & \multicolumn{5}{c}{HPSv2.1} & \multirow{2}{*}{ImageReward$\uparrow$} & \multirow{2}{*}{Latency(s)$\downarrow$} & \multirow{2}{*}{Speedup} \\
\cmidrule(lr){2-6}
& Photo & Concept-Art & Anime & Paintings & Overall$\uparrow$ &  &  &  \\
\midrule
Infinity-2B & 29.40 & 30.38 & 31.72 & 30.39 & 30.47 &  0.94 & 1.57 & 1.0 $\times$ \\
+FastVAR    & 28.86 & 29.90 & 31.12 & 29.92 & 29.95 & 0.92 & 0.62 & 2.5 $\times$ \\
+SkipVAR    & 29.25 & \textbf{30.25} & 31.50 & \textbf{30.45} & \textbf{30.39} & 0.93 & 0.87 & 1.8 $\times$ \\
\rowcolor{gray!15}
+ToProVAR   & \textbf{29.26} & 30.15 & \textbf{31.44} & 30.23 & 30.27 & \textbf{0.93} & \textbf{0.58} & \textbf{2.7} $\times$ \\
\midrule
Infinity-8B & 29.42 & 31.27 & 32.45 & 30.83 & 30.99 &  1.04 & 5.31  & 1.0 $\times$ \\
+FastVAR    & \textbf{29.87} & 30.42 & 31.80 & 29.89 & 30.24 & 1.02 & 1.97 & 2.6 $\times$ \\
+SkipVAR    & 29.09 & 30.86 & 32.04 & \textbf{30.55} & \textbf{30.64} & 1.03 & 2.65 & 2.0 $\times$ \\
\rowcolor{gray!15}
+ToProVAR   & 29.19 & \textbf{30.89} & \textbf{32.09} & 30.24 & 30.58 & \textbf{1.04} & \textbf{1.75} & \textbf{3.0 $\times$} \\
\bottomrule
\end{tabular}
\end{table}

\textbf{Models and Evaluation.} We conduct experiments on Infinity-2B and Infinity-8B\citep{han2024infinity}. We compare ToProVAR with representative approaches such as FastVAR\citep{guo2025fastvar} and SkipVAR\citep{li2025skipvar}, while keeping all hyperparameters consistent for fairness. For evaluation, we adopt widely used benchmarks including Geneval\citep{ghosh2024geneval}, DPG-Bench\citep{Hu2024Ella}, HPSv2\citep{wu2023hpsv21}, ImageReward~\citep{xu2023imagereward}, and MJHQ30K\citep{li2024mjhq}. We report Geneval Overall, DPG Overall, HPSv2 score, FID, and CLIP score as quality metrics. Efficiency is assessed in terms of runtime, throughput, and speedup ratio, with latency measured on a single NVIDIA L40 GPU (40GB). 

\begin{table}[t]

\caption{\textbf{Quantitative comparisons} of FID and CLIP score with different generation categories on the MJHQ30K dataset. Latency is measured on a single GPU with batch size $1$.}
\label{tab:adjusted_table}
\centering
\small
\setlength{\tabcolsep}{7pt}
\begin{tabular}{lccccccccc}
\toprule
\multirow{2}{*}{Method} & \multicolumn{3}{c}{Landscape} & \multicolumn{3}{c}{People} & \multicolumn{3}{c}{Food} \\
\cmidrule(lr){2-4} \cmidrule(lr){5-7} \cmidrule(lr){8-10}
& Latency & FID$\downarrow$ & CLIP$\uparrow$ & Latency & FID$\downarrow$ & CLIP$\uparrow$ & Latency & FID$\downarrow$ & CLIP$\uparrow$ \\
\midrule
Infinity-2B & 1.67 & 44.1 & 0.267 & 1.71 & 58.91 & 0.281 & 1.69 & 84.2 & 0.270 \\
+FastVAR    & 0.60 & 45.1 & 0.264 & 0.61 & 71.8 & 0.274 & 0.61 & 84.7 & 0.273 \\
+SkipVAR    & 1.01 & 58.1 & 0.260 & 1.06 & 73.7 & 0.253 & 0.88 & 102.3 & 0.256 \\
\rowcolor{gray!15}
+ToProVAR   & \textbf{0.50} & \textbf{44.5} & \textbf{0.264} & \textbf{0.48} & \textbf{58.84} & \textbf{0.283} & \textbf{0.46} & \textbf{84.3} & \textbf{0.274} \\
\bottomrule
\end{tabular}
\end{table}

\subsection{Main Result}

\textbf{Quantitative Comparison on GenEval and DPG.} We evaluate ToProVAR on GenEval and DPG to assess the quality–efficiency trade-off (Table~\ref{tab:geneval_dpg}). On Infinity-2B, ToProVAR achieves a 3.4 $\times$ speedup while maintaining the same GenEval score as the baseline, and even improves fine-grained metrics such as \emph{Position} (+5.5) and \emph{Color Attribute} (+4.0). On Infinity-8B, it delivers a 2.7 $\times$ speedup with no quality degradation, reaching the highest overall scores on both benchmarks. These results show that ToProVAR improves efficiency without sacrificing quality.

\textbf{Quantitative Comparison on HPSv2 and ImageReward.} We conduct evaluations on the HPSv2.1 and ImageReward benchmarks to comprehensively evaluate the human preference performance. As shown in Table~\ref{tab:hpsv2_imagereward}, ToProVAR achieves significant acceleration while maintaining high-quality generation. On the Infinity-2B model, ToProVAR reduces inference latency by 62.4\% with a negligible drop in the overall HPSv2 score (\textless 1\%). The advantages of ToProVAR are even more pronounced on the larger Infinity-8B model, where it reduces latency by 67\%  while preserving the same ImageReward score and a highly competitive HPSv2 score. These results powerfully demonstrate that ToProVAR strikes an exceptional balance between efficiency and quality, effectively upholding subjective image quality and human preference alignment while improving inference speed.

\begin{figure}[t]
    \includegraphics[width=\linewidth]{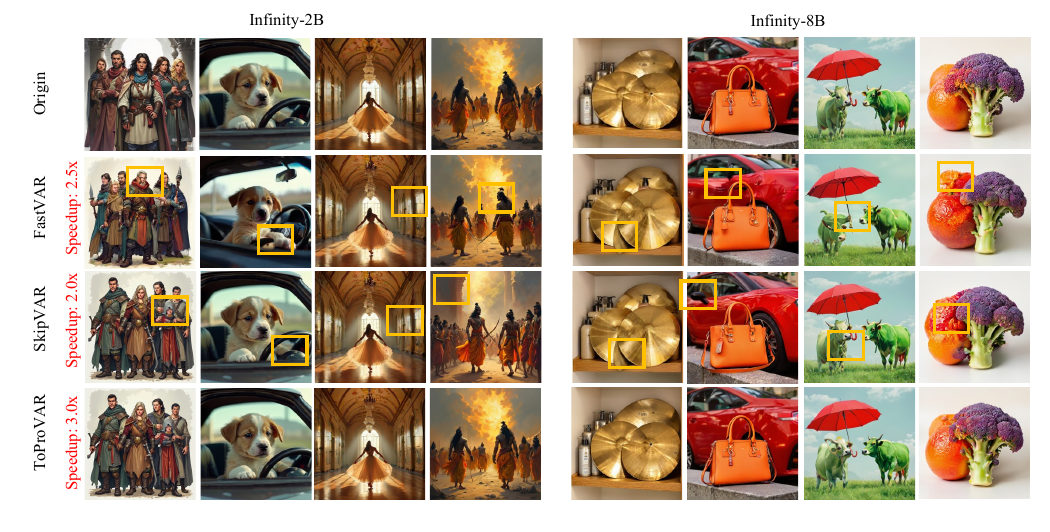}
    \vspace{-7mm}
    \caption{\textbf{Qualitative comparison} of various methods on complex prompts. Our method effectively prevents semantic loss, structure distortion, and detail collapse while maintaining high visual
fidelity. }
    \label{fig:5}
\end{figure}

\begin{table}[t]
    \centering
    \begin{minipage}{0.55\textwidth}
        \setlength{\tabcolsep}{2pt}
        \caption{Ablation study of the Three-Dimensional Progressive Manipulation Framework on Infinity-2B, where ``+'', ``++'', and ``+++'' denote progressive component addition over the previous stage.}
        \label{tab:components}
        \centering
        \scalebox{0.85}{
        \begin{tabular}{lccc}
        \toprule
        Method & Latency(s)$\downarrow$ & Speed $\uparrow$ & GenEval$\uparrow$ \\
        \midrule
        Infinity-2B &2.10  & - & 0.690 \\
        + Scale Depth Loc. & 0.47 & 4.5 $\times$ & 0.477 \\
        ++ Layer Repr. Ident. &0.57  & 3.7 $\times$ & 0.679 \\
        +++ Fine-grained Token Prun. & 0.61 & 3.4 $\times$ & 0.690 \\
        \bottomrule
        \end{tabular}}
    \end{minipage}
    \hfill
    \begin{minipage}{0.44\textwidth}
        \setlength{\tabcolsep}{4pt}
        \caption{Ablation Studies of Flash Attention Entropy on Infinity-2B.``w/o FAE'' denotes naive attention entropy calculation, which creates computational overhead.}
        \label{tab:function}
        \centering
        \scalebox{0.85}{
        \begin{tabular}{lcc}
        \toprule
        Setup & Speed$\uparrow$ & Latency(s)$\downarrow$ \\
        \midrule
        Infinity-2B & - & 2.10 \\
        +FastVAR & 2.6 $\times$ & 0.80 \\
        +ToProVAR(w/o FAE) & 1.9 $\times$ & 1.10  \\
        +ToProVAR(w/ FAE) & 3.4 $\times$ & 0.61 \\
        \bottomrule
        \end{tabular}}
    \end{minipage}
\end{table}

\begin{figure*}[t]
    \includegraphics[width=\linewidth]{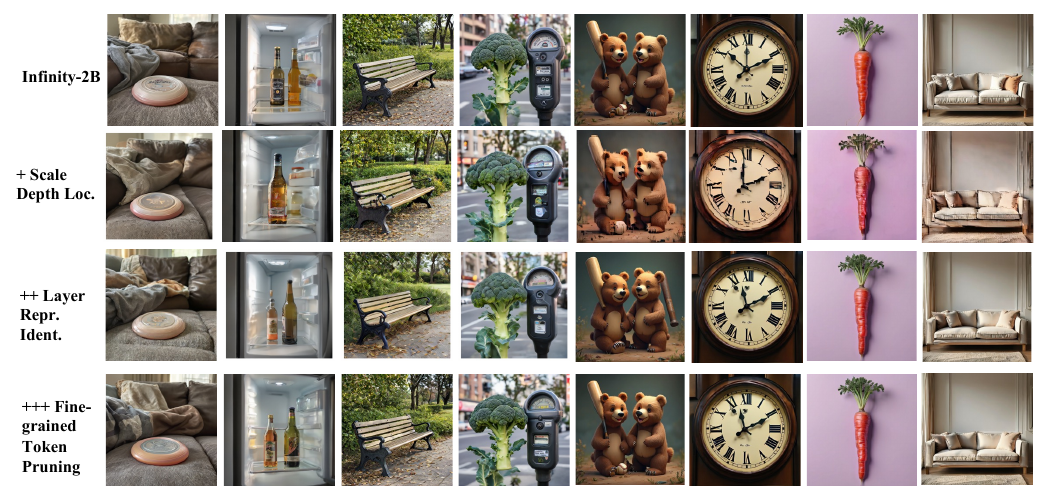}
    \caption{Qualitative Visualizations of the three components ablation(Scale, Layer, Token) on Infinity-2B.The full tri-stage framework better preserves global layout and local details than partial variants that prune along only a subset of dimensions.}
    \label{fig:6}
\end{figure*}

\textbf{Quantitative Comparison on MJHQ30K.} In Table~\ref{tab:adjusted_table}, we validate the perceptual quality on the MJHQ30K benchmark. It can be seen that our ToProVAR achieves a reasonable performance while maintaining a high speedup ratio. For instance, on the challenging People category, our ToProVAR even achieves a FID reduction with 3.5$\times$ acceleration, decreasing FID from 58.91 to 58.84. In other categories, our method also maintains good performance with significant acceleration.

\textbf{Qualitative Visualizations of Different Methods.}
Figure~\ref{fig:5} presents qualitative results of different methods on Infinity-2B/8B. Compared to FastVAR and SkipVAR, ToProVAR effectively mitigates semantic loss, structural distortion, and detail collapse. It maintains high visual fidelity while achieving a greater acceleration ratio. 

\subsection{Ablation Study}

\textbf{Impact of Stages in the Tri-Dimensional Optimization Framework.}
To validate the effectiveness of our design, we analyze the contribution of each stage in the framework in Table~\ref{tab:components} and Figure~\ref{fig:6}. A coarse-grained approach using only Scale Depth Localization and Layer Representation Identification is fast but suboptimal, with a GenEval score of 0.679. Incorporating Fine-grained Token Pruning completes the ToProVAR model, restoring the score to 0.690—nearly matching the baseline—while maintaining a low 0.61s latency. This result highlights that our complete tri-stage framework is essential for optimally balancing acceleration and fidelity.

\textbf{Ablation Studies on Flash Attention Entropy.}
We validate the efficiency of our integrated Flash Attention Entropy (FAE) module in Table~\ref{tab:function}. Naively attention entropy calculation without FAE creates a significant computational bottleneck, increasing latency to 1.10s. In contrast, our fully integrated module eliminates this overhead and reduces latency to 0.61s. This result confirms that FAE is critical for ToProVAR, as it preserves and enhances the acceleration benefits of Flash Attention.

\begin{figure}[t]
    \centering
    \begin{minipage}[t]{0.49\textwidth}
        \centering
        \vspace{0pt}
        \setlength{\tabcolsep}{5pt}
        \captionof{table}{Time cost of frequency-, entropy-, and SVD-related operations on Infinity-8B.}
        \vspace{-8pt} 
        \label{tab:6}
        \scalebox{0.85}{
        \begin{tabular}{lcc}
            \toprule
            \textbf{Operation / Time (ms)} & \textbf{All Scales} & \textbf{Rep. Scale s} \\
            \midrule
            \multicolumn{3}{l}{\textit{Frequency \& Entropy ($s=10$)}} \\
            \midrule
            Frequency-based scoring             & 1.30  & 0.16 \\
            Attention Entropy (na\"{i}ve)       & 125.73 & 12.06 \\
            FlashAttention                      & 11.27 & 1.11 \\
            Flash Attention Entropy & 12.97 & 1.28 \\
            \midrule
            \multicolumn{3}{l}{\textit{Layer-level SVD ($s=6$)}} \\
            \midrule
            SVD per layer                       & --    & 1.25 \\
            SVD over all layers                 & --    & 49.84 \\
            \bottomrule
        \end{tabular}
        }
    \end{minipage}
    \hfill
    \begin{minipage}[t]{0.49\textwidth}
        \centering
        \vspace{0pt}
        \includegraphics[width=\textwidth]{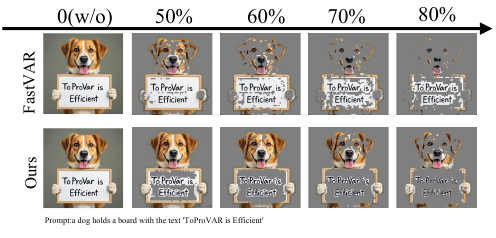}
        \captionof{figure}{Visualization of pruned tokens by FastVAR and ToProVAR.}
        \label{fig:7}
    \end{minipage}
\end{figure}

\textbf{Visualization of pruned tokens with different methods.}
Fig.~\ref{fig:7} visualizes the divergent pruning strategies of FastVAR and ToProVAR. While FastVAR's frequency-based heuristic retains edges, it erodes an image's semantic integrity by removing tokens from smooth yet vital areas like facial contours. This structural damage becomes severe as the pruning ratio increases. In contrast, ToProVAR leverages attention entropy to identify and preserve core semantic content. Even at a 90\% pruning ratio, it protects critical features like the eyes. This visu  al comparison confirms that attention entropy is a more robust heuristic than frequency for preserving semantic structure during pruning.

\textbf{Computational Cost Analysis.}
We further quantify the overhead of our tri-dimensional sparsity framework, focusing on Flash Attention Entropy (FAE) and the layer-level SVD analysis.
As shown in Table~\ref{tab:6}, na\"{i}ve attention entropy incurs a severe bottleneck due to explicit attention-matrix materialization, whereas FAE computes entropy on-the-fly inside the FlashAttention kernel with only a lightweight $x\log x$ reduction. This yields only 0.17\,ms overhead over FlashAttention at scale 10, yet reduces entropy cost by \(\sim\)90\%.
For layer analysis, SVD is performed once per layer at a single representative scale ($s{=}6$). Table~\ref{tab:6} shows 49.84\,ms total over all layers, i.e., $<3\%$ of the 1780\,ms end-to-end latency, indicating minor overhead.

\section{Related Work}
\paragraph{Autoregressive Visual Generation.} 
AR models~\citep{li2024mar, liu2024lumina, SandAI2025MAGI1, xie2024showo}), originally successful in language, have been extended to image generation through next-token prediction~\citep{van2017vqvae, Dai_2025_ICCV}.
Recent scaling has narrowed the gap with diffusion models~\citep{xie2024showo,wu2024vila,wang2024emu3,wu2024janus}.
Yet efficiency remains a bottleneck. 
To address this, Visual Autoregressive (VAR) modeling\citep{tian2024var,han2024infinity, tang2024hart} introduces a next-scale paradigm, where images are predicted in hierarchical token maps from coarse to fine resolution. 
This strategy reduces the number of autoregressive steps and improves both speed and quality.

\paragraph{Efficient Visual Generation.}
The acceleration of diffusion models has been extensively studied, with mature methods including distillation~\citep{Kim_2025_CVPR,zhai2024motion,yin2025causvid}, quantization~\citep{zhao2024viditq, xi2025sparse, wu2025quantcache}, pruning~\citep{zou2024accelerating, Fang_2025_CVPR, 10946713}, and feature caching~\citep{Liu_2025_CVPR,zhao2024real,ma2023deepcache,lv2024fastercache,liu2025taylorseer}. 
Efficient though, they are tailored to diffusion architectures and cannot be directly applied to the hierarchical prediction in VAR.

Efficiency optimization for VAR is still nascent. Early works such as FastVAR~\citep{guo2025fastvar} and SkipVAR~\citep{li2025skipvar} exploit fixed pruning or frequency-based skipping, while recent methods explore semantic- and structure-aware acceleration, e.g., SparseVAR~\citep{chen2025frequencyawareautoregressivemodelingefficient} for token sparsity and CoDe~\citep{Chen_2025_CVPR} for collaborative decoding. Complementary insights are provided by methods such as HACK~\citep{qin2025headawarekvcachecompression} and ScaleKV~\citep{li2025scalekv}, which focus on KV cache compression.

\section{Conclusion}

In this work, we present \textbf{ToProVAR}, a novel acceleration framework that addresses tri-dimensional redundancies of Visual Autoregressive models. 
\textbf{ToProVAR} leverages attention entropy to uncover sparsity patterns across tokens, layers, and scales.
With fine-grained semantic modeling and pruning strategy, critical contents are preserved even with aggressive acceleration, achieving 3.4$\times$ speedup with minimal quality loss, surpassing existing methods. 
Our study highlights the value of semantic-driven optimization for AR generation and future extensions in video and multimodal modeling.

\clearpage

\bibliography{_ref/3_VAR}
\bibliographystyle{iclr2026_conference}

\clearpage

\newcommand{\vQ}{\mathbf{Q}}
\newcommand{\vK}{\mathbf{K}}
\newcommand{\vV}{\mathbf{V}}
\newcommand{\vdQ}{\mathbf{dQ}}
\newcommand{\vdK}{\mathbf{dK}}
\newcommand{\vdV}{\mathbf{dV}}
\newcommand{\vS}{\mathbf{S}}
\newcommand{\vdS}{\mathbf{dS}}
\newcommand{\vP}{\mathbf{P}}
\newcommand{\vdP}{\mathbf{dP}}
\newcommand{\vU}{\mathbf{U}}
\newcommand{\vW}{\mathbf{W}}
\newcommand{\vT}{\mathbf{T}}
\newcommand{\vX}{\mathbf{X}}
\newcommand{\vO}{\mathbf{O}}
\newcommand{\vdO}{\mathbf{dO}}
\newcommand{\vM}{\mathbf{M}}
\newcommand{\vZ}{\mathbf{Z}}
\newcommand{\diag}{\mathrm{diag}}

\appendix
\section{Appendix}

\subsection{Experiment Details}

\label{sec:a3}

\subsubsection{Models}
Our evaluation is based on the state-of-the-art Visual Autoregressive Models, specifically Infinity-2B and Infinity-8B\citep{han2024infinity}. These models have demonstrated exceptional performance across a wide range of image generation tasks. For our experiments, we utilize their pre-trained versions and adopt their default inference configurations, with the prime structures detailed in Table \ref{tab:models}.

\begin{table}[htbp]
  \centering
  \caption{The basic information of models.}
  \label{tab:models}
  \begin{tabular}{ccc}
    \toprule
    model  & scales & layers\\
    \midrule
    Infinity-2B  & 11 & 32 \\
    Infinity-8B  & 13  & 40\\
    \bottomrule
  \end{tabular}
\end{table}

\subsubsection{Baselines Settings}
To ensure a fair comparison, we standardized experimental parameters, such as the random seed, across all compared models—Infinity, FastVAR\citep{guo2025fastvar}, SkipVAR\citep{li2025skipvar}, and our own ToProVAR—to eliminate the influence of confounding variables.

We empirically analyzed the hyperparameters of FastVAR and found that its 32 ratio and 40 ratio parameters had a non-monotonic, convex-like effect on generation quality. Consequently, we adopted the optimal, default parameter configuration for our comparisons. For SkipVAR, we directly used its default decision model of (0.84). The specific parameter settings are detailed in the relevant tables.

\begin{table}[h!]
    \centering
    \caption{Acceleration Configurations for FastVAR and SkipVAR. NOTE: Notation 's:p' means scale index s is pruned with pruning ratio p,All ratios are applied per-layer at inference time.}
    \label{tab:acceleration_config}
    \begin{tabular}{lcc}
        \toprule
        \textbf{Method} & \textbf{Model} & \textbf{Method Parameter} \\
        \midrule
        FastVAR & Infinity-2B & \{10:0.4;11:0.6\}  \\
        & Infinity-8B & \{10:0.4;11:0.6;12:1.0;13:1.0\}  \\
        \midrule
        SkipVAR & Infinity-2B & 0.84 \\
        & Infinity-8B & 0.84@2B  \\
        \bottomrule
    \end{tabular}
\end{table}

\subsubsection{Metrics}

In this work, we employ a diverse set of established metrics to comprehensively evaluate our method, aiming to assess both objective image generation quality and adherence to human instructions.

First, we utilize Geneval\citep{ghosh2024geneval} and DPG-bench\citep{Hu2024Ella} to focus on the objective quantification of generation quality.

\textbf{Geneval.} Geneval serves as a crucial tool for measuring foundational generation quality. It decomposes the task into six fine-grained sub-tasks: single-object generation, object co-occurrence, counting, color control, relative positioning, and attribute binding. By using a pre-trained detector to compare generated results with ground-truth annotations, this metric outputs multi-dimensional compliance scores, and its average serves as a comprehensive quality measure.

\textbf{DPG-bench.} DPG-bench is a specialized evaluation benchmark for scenarios involving dense prompts. It focuses on the generation quality of multi-object, multi-attribute, and multi-relational descriptions, serving as a key indicator of a model’s ability to align with complex semantics and follow instructions.

Second, we leverage HPSv2\citep{wu2023hpsv21}, ImageReward\citep{xu2023imagereward}, and SSIM\citep{wang2004ssim} to establish a quantitative link between our generated results and human perception, focusing on perceptual similarity and visual preference.

\textbf{Human Preference Score v2(HPSv2).} HPSv2 is designed to measure the alignment between generated content and human hierarchical visual perception. It uses a pre-trained visual network to extract both "low-level features (edges, colors)" and "high-level features (object structure, semantics)" from the generated and reference content, then aggregates them to yield a final score. This metric effectively gauges the perceptual plausibility of the generated content from a human perspective.

\textbf{ImageReward(IR).} IR is another important metric for aligning text-to-image generation with human preferences. It directly outputs a preference score, quantifying the visual appeal and realism of the generated content.

\textbf{Structural Similarity Index Measure (SSIM).} To facilitate the analysis of generation states across images of varying complexity, we introduce SSIM. It provides a measure of image quality that reflects structural and perceptual differences. The formula is defined as:
$$
\text{SSIM}(x, y) = \frac{(2\mu_x\mu_y + C_1)(2\sigma_{xy} + C_2)}{(\mu_x^2 + \mu_y^2 + C_1)(\sigma_x^2 + \sigma_y^2 + C_2)}
$$

\subsubsection{Generalization Experiments}
To assess the generalization of ToProVAR beyond the Infinity series, we further evaluate it on HART~\cite{tang2024hart}, a VAR-style hybrid autoregressive transformer.For HART, we use the official pre-trained checkpoint and default sampling configuration.
The resulting quality–efficiency comparison between HART, HART+FastVAR, and HART+ToProVAR is reported in Table~\ref{tab:hart_results}, showing that ToProVAR maintains comparable GenEval scores to the HART baseline while achieving additional speedups and a better quality–efficiency trade-off than FastVAR.

\begin{table}[h!]
    \centering
    \caption{Comparison of ToProVAR and FastVAR on the GenEval benchmark using HART.}
    \label{tab:hart_results}
    \small
    \begin{tabular}{lccccccc}
        \toprule
        \textbf{Method} & \textbf{Two Obj.} & \textbf{Position} & \textbf{Color} & \textbf{Attri.} & \textbf{Overall} $\uparrow$ & \textbf{Latency (s)} $\downarrow$ & \textbf{Speedup} $\uparrow$ \\
        \midrule
        HART            & 0.62 & 0.13 & 0.18 & 0.51 & 0.51 & 0.95 & 1.0$\times$ \\
        +FastVAR        & 0.59 & 0.13 & 0.19 & 0.50 & 0.50 & 0.64 & 1.5$\times$ \\
        +ToProVAR (ours)& 0.61 & 0.13 & 0.18 & 0.51 & 0.51 & 0.56 & 1.7$\times$ \\
        \bottomrule
    \end{tabular}
\end{table}

\subsubsection{Further Tri-Dimensional Ablation Experiments}

In the Table~\ref{tab:components} of main paper, we provide a three-stage ablation on Infinity-2B.
To more clearly isolate the contributions of each dimension in our tri-dimensional framework
(Scale $\rightarrow$ Layer $\rightarrow$ Token), we further introduce two controlled variants and conduct an extended ablation, summarized in Table~\ref{tab:all_components}.

Concretely, all variants share the same Infinity-2B backbone and sampling settings, and differ only in
which sparsity dimensions are activated:

\begin{itemize}
    \item \textbf{Fix Scale Exit.} A coarse baseline that skips a fixed set of late scales with a
    hand-crafted exit scale, without any Scale Depth Localization. This variant achieves the highest
    nominal speed but suffers from the most severe quality degradation, highlighting the importance
    of \emph{adaptive} scale selection.

    \item \textbf{Scale Depth Localization (Scale).} A pure scale-skipping variant in which only the
    Scale Depth Localization module is enabled. It adaptively selects the pruning start scale and
    skips subsequent scales, improving efficiency over the baseline, but may still introduce
    noticeable semantic distortions due to the lack of layer- and token-level control.

    \item \textbf{Scale Depth Loc. + Fine-grained Token Pruning (Scale + Token).} A configuration that
    combines adaptive scale skipping with token-level pruning applied uniformly across \emph{all}
    layers, without Layer Representation Identification. While this improves over purely scale-level
    pruning, its quality remains clearly below the baseline, indicating that unconstrained token
    pruning on Global layers can harm global semantics.

    \item \textbf{Scale Depth Loc. + Layer Representation Identification (Scale + Layer).} A
    scale–layer variant in which we perform adaptive scale skipping together with layer skipping
    guided by the layer-scope analysis. Layers identified as Detail are skipped more aggressively,
    whereas Global layers are largely preserved. This substantially recovers global structure and
    semantics while maintaining strong acceleration.

    \item \textbf{Full ToProVAR (Scale + Layer + Token).} The complete tri-stage framework, which
    combines adaptive scale skipping, layer skipping based on Global/Detail classification, and
    fine-grained token pruning restricted to selected Detail layers. This configuration restores
    generation quality to be on par with the baseline while retaining substantial speedup.
\end{itemize}

Overall, these extended ablations demonstrate the \emph{progressive} and \emph{complementary} roles of
scale-, layer-, and token-level optimization in balancing acceleration and fidelity: naive scale-only
or scale+token pruning can be overly aggressive, whereas the full tri-dimensional design of ToProVAR
achieves a much better quality–efficiency trade-off. These configurations correspond to the rows in
Table~\ref{tab:all_components} and the visual comparisons in Figure~\ref{fig:6}, and are used consistently across all ablation experiments.

\begin{table}[h!]
    \centering
    \caption{Extended ablation study of the tri-dimensional progressive framework on Infinity-2B.}
    \label{tab:all_components}
    \begin{tabular}{lccc}
        \toprule
        \textbf{Method} & \textbf{Latency (s)} $\downarrow$ & \textbf{Speed} $\uparrow$ & \textbf{GenEval} $\uparrow$ \\
        \midrule
        Infinity-2B                                  & 2.10 & 1.0$\times$ & 0.690 \\
        Fix Scale Exit                               & 0.41 & 5.1$\times$ & 0.403 \\
        Scale Depth Loc.                             & 0.47 & 4.5$\times$ & 0.477 \\
        Scale Depth Loc. + Fine-grained Token Prun.  & 0.78 & 2.7$\times$ & 0.603 \\
        Scale Depth Loc. + Layer Repr. Ident.        & 0.57 & 3.7$\times$ & 0.679 \\
        ToProVAR (Scale + Layer + Token)             & 0.61 & 3.4$\times$ & 0.690 \\
        \bottomrule
    \end{tabular}
\end{table}

\begin{figure}[t]
    \includegraphics[width=\linewidth]{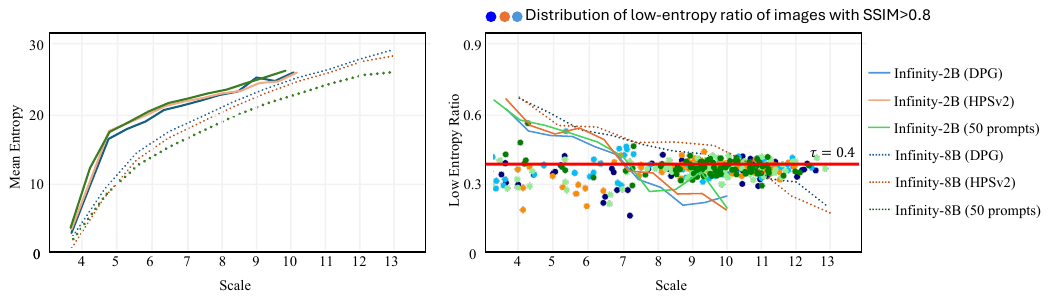}
    \caption{Robustness analysis of the scale-depth threshold $\tau$. Left: mean attention entropy vs. scale. Right: low-entropy ratio $\rho_s$ vs. scale, together with the distribution of $\rho_s$ for images whose SSIM to the full-scale baseline exceeds $0.8$.}
    \label{fig:robust}
\end{figure}

\subsection{Calibration and robustness of the scale-depth threshold $\tau$}
\label{sec:appendix_tau}

Building on the scale-level analysis in Eq.~(3), we quantify the semantic fineness at scale $s$ by the
low-entropy ratio
\begin{equation}
    \rho_s = \frac{\lvert \{ i \mid H_i^s < \bar{H}^s \} \rvert}{N_s},
\end{equation}
where $H_i^s$ denotes the attention entropy of token $i$ at scale $s$, $\bar{H}^s$ is the mean entropy
at that scale, and $N_s$ is the number of tokens.

To calibrate the pruning start depth, we perform a lightweight pre-sampling procedure on a small set of
prompts and compute statistics across scales. For each backbone (e.g., Infinity-2B, Infinity-8B), we
randomly sample a calibration subset of prompts from HPSv2 and DPG-Bench, and plot both the
\emph{Mean-Entropy–scale} curves and the \emph{$\rho_s$–scale} curves, as shown in
Fig.~\ref{fig:robust}. We obtain the following empirical observations:

\begin{itemize}
    \item \textbf{Consistency of mean entropy across datasets.}  
    As shown in the left panel, for a fixed backbone, the mean attention entropy at each scale is
    highly consistent across HPSv2, DPG-Bench, and a randomly sampled 50-prompt subset: both the
    absolute values and the scale-wise trends of the curves almost overlap. This indicates that the
    attention-entropy statistics are largely insensitive to the specific dataset or calibration subset.
    
    \item \textbf{Stability of the low-entropy ratio across scales.}  
    The right panel shows that the $\rho_s$–scale curves for different datasets and calibration-set
    sizes (50 prompts vs. the full prompt set) exhibit very similar trajectories. For a given backbone,
    the scale index at which $\rho_s$ enters a “reasonable” band is highly stable across datasets and
    prompt subsets.

    \item \textbf{Low-entropy ratio as a quality indicator and choice of $\tau = 0.4$.}  
    The scatter points in the right panel correspond to images whose SSIM with respect to the
    full-scale baseline is greater than $0.8$. These high-quality generations concentrate around a
    narrow range of low-entropy ratios, approximately centered at $\rho_s \approx 0.4$. This
    observation suggests that $\rho_s$ can serve as a proxy for generation quality, and motivates the
    choice of $\tau = 0.4$ as a quality-aware threshold.
\end{itemize}

Based on this invariance, we choose a single global threshold $\tau$ per backbone such that
$\rho_s \ge \tau$ marks the onset of semantically stable scales suitable for pruning. In practice,
$\tau$ is calibrated once on the small calibration subset and then reused across all datasets,
resolutions, and prompts, without any dataset-specific hyperparameter search. This explains why
the same $\tau$ generalizes well in all our experiments and why our scale-level pruning remains
robust under variations in prompts and resolutions.

\subsection{More Visualization Results}

This section presents additional visualizations that complement the observations described in the main text.

\begin{figure*}[t]
    \includegraphics[width=\linewidth]{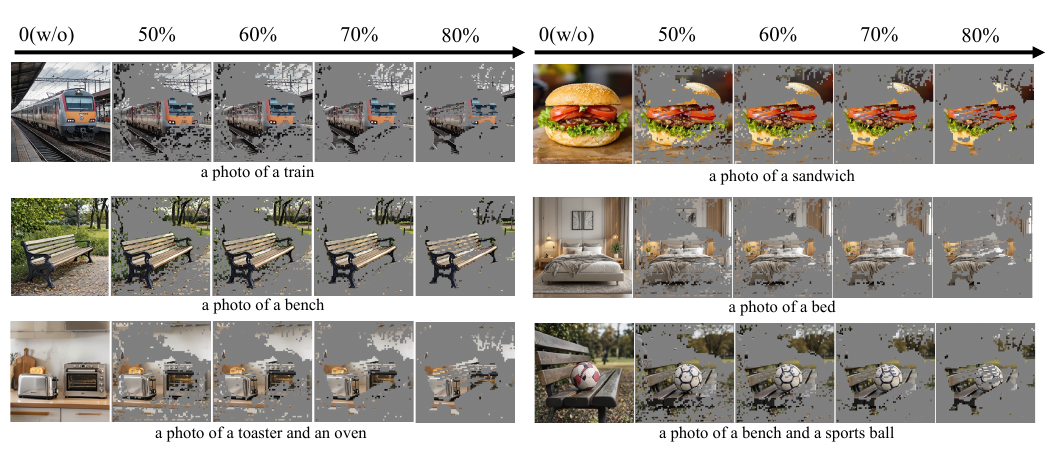}
    \vspace{-7mm}
    \caption{Visualization of token-level pruning in ToProVAR. Gray tokens indicate those pruned by ToProVAR, while colored tokens correspond to preserved, semantically salient regions. }
    \label{fig:token_vis_more}
\end{figure*}

\subsubsection{Pruned Tokens Visualizations}
\label{sec:appendix_token}

In this section, we visualize the token-level pruning decisions made by ToProVAR on image generation
tasks with varying levels of complexity. For each example, tokens that are \emph{pruned} by ToProVAR are rendered in gray, while \emph{preserved} tokens retain their original image appearance.
As shown in Figure~\ref{fig:token_vis_more}, ToProVAR primarily removes tokens in redundant background
regions and keeps tokens concentrated around object contours and fine details, leading to more accurate and semantically aligned sparsity patterns.


\begin{figure*}[t]
    \centering
    \includegraphics[width=\linewidth]{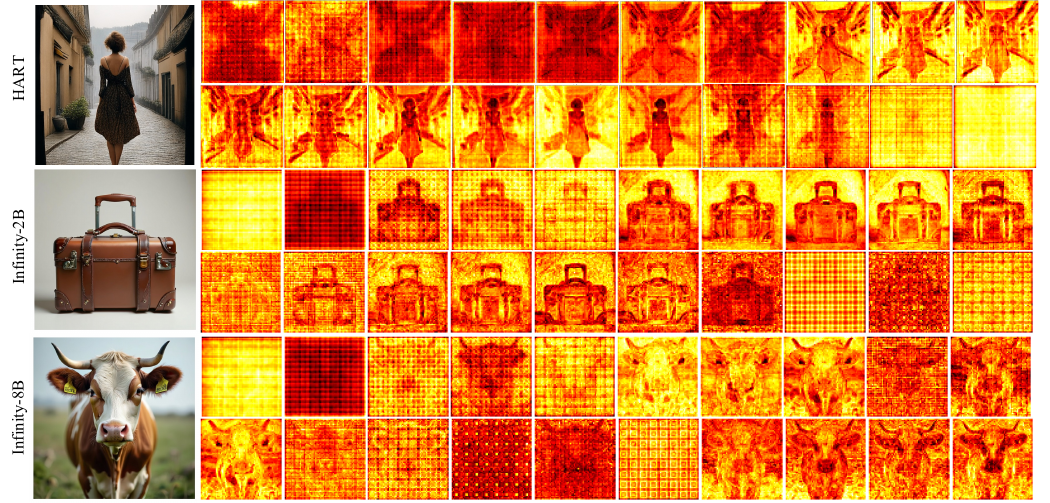}
    \caption{Additional visualizations of layer representations on HART, Infinity-2B, and Infinity-8B.We show example attention maps for the representative Global and Detail layers. }
    \label{fig:layer_1}
\end{figure*}

\begin{figure*}[t]
    \centering
    \includegraphics[width=\linewidth]{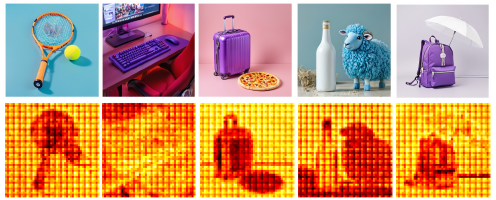}
    \caption{Additional visualizations of failure layer representation case. }
    \label{fig:layer_failure}
\end{figure*}

\subsubsection{Layer-Level Semantic Representation Visualizations}
\label{sec:appendix_layer}

In this section, we provide additional visualizations of layer-level semantic representations across different VAR backbones. 
As shown in Figure~\ref{fig:layer_1}, we consistently observe two dominant patterns along the layer dimension: some layers exhibit grid-like, globally distributed attention,
while others focus on localized, fine-grained regions. This dichotomy underpins our strategy of categorizing layers into Global and Detail layers, and it substantiates the design of Stage~II, where semantic analysis and pruning are performed at the layer level.

Crucially, this behavior is not restricted to the Infinity series. When we apply the same layer-scope analysis to HART, we find a very similar organization: early and final layers tend to act as Global Layers with a dominant principal component, whereas middle layers behave as Detail Layers with more localized and diverse semantics. This pattern is stable across prompts and scales when we classify layers at a representative, semantically stable scale.

Figure~\ref{fig:layer_failure} further illustrates the rare \emph{ambiguous} cases. In these layers, the attention maps exhibit both weak grid-like global structure and pronounced object-centric activations, so that the layer plays a mixed role of refining global layout and local details. Such hybrid behavior typically arises at transition layers where global composition is being finalized while fine details start to emerge, and is further amplified by averaging over heads, since different heads may specialize in global versus local semantics. Consequently, these layers lie close to the decision boundary of our principal-component–based classifier and can be labeled as Global or Detail depending on small variations across prompts or scales. However, they account for only a very small subset of layers we inspected, and their impact on final accuracy is negligible: our pruning policy is conservative on Detail layers, and the semantics captured by these ambiguous layers are largely redundant with neighboring layers.

\subsubsection{Qualitative Comparison of Various Methods}
\label{sec:appendix_methods}

We further present qualitative comparisons of FastVAR, SkipVAR, and ToProVAR on the Infinity-2B and Infinity-8B backbones. As shown in Figure~\ref{fig:7}, we visualize generated images across diverse prompts and scenes, which allows an intuitive assessment of the quality–efficiency trade-offs achieved by each method. In particular, ToProVAR tends to better preserve global layout and fine-grained details
while operating at comparable or higher acceleration levels.

\begin{figure*}[t]
    \includegraphics[width=\linewidth]{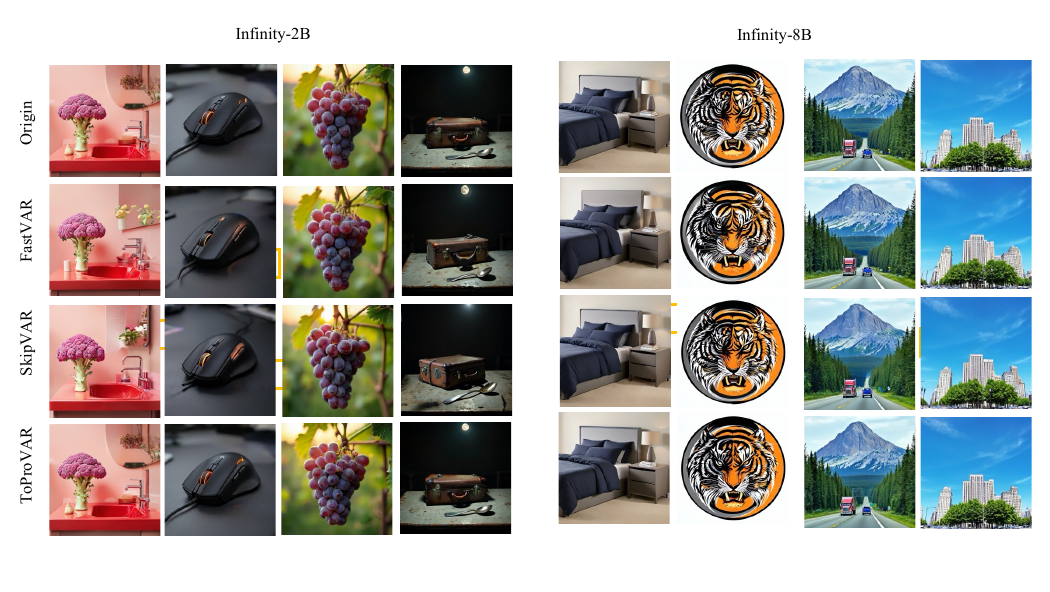}
    \vspace{-7mm}
    \caption{Additional visualization results of FastVAR, SkipVAR, and ToProVAR on Infinity-2B and Infinity-8B.
    Compared to the baselines, ToProVAR more faithfully preserves global structure and fine details under
    similar or higher acceleration, yielding visually sharper and more semantically consistent generations.}
    \label{fig:7}
\end{figure*}

\subsection{Flash Attention Entropy (FAE)}
\label{sec:a5}

\begin{algorithm}[b]
  \baselineskip=0.7em
  \caption{\small\label{alg:flash2_fwd}original FlashAttention forward pass}
  \begin{algorithmic}[1]
    \REQUIRE Matrices $\vQ, \vK, \vV \in \mathbb{R}^{N \times d}$ in HBM, block sizes $B_c$, $B_r$.
    \STATE \label{alg:stream_attn_split_qkv} Divide $\vQ$ into $T_r = \left\lceil\frac{N}{B_r} \right\rceil$ blocks $\vQ_1, \dots, \vQ_{T_r}$ of size $B_r \times d$ each,
    and divide $\vK, \vV$ in to $T_c = \left\lceil \frac{N}{B_c} \right\rceil$ blocks $\vK_1, \dots, \vK_{T_c}$ and
    $\vV_1, \dots, \vV_{T_c}$, of size $B_c \times d$ each.
    \STATE Divide the output $\vO \in \mathbb{R}^{N \times d}$ into $T_r$ blocks $\vO_i, \dots, \vO_{T_r}$ of size
    $B_r \times d$ each, and divide the logsumexp $L$ into $T_r$ blocks $L_i, \dots, L_{T_r}$ of size
    $B_r$ each.
    \FOR{$1 \le i \le T_r$} \label{alg:stream_attn_outer_loop}
      \STATE \label{alg:stream_attn_load_q} Load $\vQ_i$ from HBM to on-chip SRAM.
      \STATE \label{alg:stream_attn_init} On chip, initialize $\vO_{i}^{(0)} = (0)_{B_r \times d} \in \mathbb{R}^{B_r \times d}, \ell_{i}^{(0)} = (0)_{B_r} \in \mathbb{R}^{B_r}, m_{i}^{(0)} = (-\infty)_{B_r} \in \mathbb{R}^{B_r}$.
      \FOR{$1 \le j \le T_c$}
        \STATE \label{alg:stream_attn_load_kv} Load $\vK_j, \vV_j$ from HBM to on-chip SRAM.
        \STATE \label{alg:stream_attn_qk} On chip, compute $\vS_{i}^{(j)} = \vQ_i \vK_j^T \in \mathbb{R}^{B_r \times B_c}$.
        \STATE \label{alg:stream_attn_statistics} On chip, compute
        $m_{i}^{(j)} = \mathrm{max}(m_{i}^{(j-1)}, \mathrm{rowmax}(\vS_{i}^{(j)})) \in \mathbb{R}^{B_r}$, $\tilde{\vP}_{i}^{(j)} = \exp(\vS_{i}^{(j)} - m_{i}^{(j)}) \in \mathbb{R}^{B_r \times B_c}$ (pointwise),
        $\ell_{i}^{(j)} = e^{m_{i}^{j-1} - m_{i}^{(j)}} \ell_{i}^{(j-1)} + \mathrm{row sum}(\tilde{\vP}_{i}^{(j)}) \in \mathbb{R}^{B_r}$.
        \STATE \label{alg:stream_attn_update} On chip, compute
        $\vO_{i}^{(j)} = \diag(e^{m_{i}^{(j-1)} - m_{i}^{(j)}})^{-1} \vO_{i}^{(j-1)} + \tilde{\vP}_{i}^{(j)} \vV_j$.
      \ENDFOR
      \STATE On chip, compute $\vO_{i} = \diag(\ell_{i}^{(T_c)})^{-1} \vO_{i}^{(T_c)}$.
      \STATE On chip, compute $L_{i} = m_{i}^{(T_c)} + \log(\ell_i^{(T_c)})$.
      \STATE Write $\vO_{i}$ to HBM as the $i$-th block of $\vO$.
      \STATE Write $L_{i}$ to HBM as the $i$-th block of $L$.
    \ENDFOR
    \STATE Return the output $\vO$ and the logsumexp $L$.
  \end{algorithmic}
\end{algorithm}

To efficiently obtain attention entropy without materializing the full attention matrix, we extend the
original FlashAttention forward pass to compute entropy on the fly inside the streaming kernel.

As summarized in Algorithm~\ref{alg:flash2_fwd}, the standard FlashAttention implementation processes
$\vQ, \vK, \vV$ in blocks, incrementally updating the partial output $\vO_i^{(j)}$ and the
normalization statistics $(m_i^{(j)}, \ell_i^{(j)})$ for each query block. At the end of the loop
over key/value blocks, it returns the normalized output $\vO$ together with the per-row log-sum-exp
$L$, which is typically used for numerical stability and backward computation.

Our Flash Attention Entropy (FAE), shown in Algorithm~\ref{alg:flash2_entropy}, augments this kernel
with an additional accumulator $E_i^{(j)}$ that maintains the running sum of $p \log p$ in the same
numerically stable streaming fashion used for $\ell_i^{(j)}$. Concretely, we reuse the intermediate
unnormalized probabilities $\tilde{\vP}_i^{(j)}$ and apply a lightweight \texttt{row\_reduce\_xlogx}
operation at each step. After processing all key/value blocks, we obtain the per-query entropy vector
$E_i$ by combining $E_i^{(T_c)}$ and $\ell_i^{(T_c)}$, in analogy to how $L_i$ is derived from
$m_i^{(T_c)}$ and $\ell_i^{(T_c)}$. This design keeps the memory footprint identical to standard
FlashAttention and avoids constructing the full $N \times N$ attention matrix.

Since our sparsity framework only requires the attention output $\vO$ and the corresponding entropy
$E$ at inference time, we do not use the log-sum-exp $L$ returned by the original kernel. In our
implementation, we therefore drop $L$ from the return values and only expose $(\vO, E)$, while the
core FlashAttention streaming structure remains unchanged.

\begin{algorithm}[t]
  \baselineskip=0.7em
  \caption{\small\label{alg:flash2_entropy}FlashAttention forward pass with entropy}
  \begin{algorithmic}[1]
    \REQUIRE Matrices $\vQ, \vK, \vV \in \mathbb{R}^{N \times d}$ in HBM, block sizes $B_c$, $B_r$.
    \STATE \label{alg:stream_attn_split_qkv} Divide $\vQ$ into $T_r = \left\lceil\frac{N}{B_r} \right\rceil$ blocks $\vQ_1, \dots, \vQ_{T_r}$ of size $B_r \times d$ each,
    and divide $\vK, \vV$ in to $T_c = \left\lceil \frac{N}{B_c} \right\rceil$ blocks $\vK_1, \dots, \vK_{T_c}$ and
    $\vV_1, \dots, \vV_{T_c}$, of size $B_c \times d$ each.
    \STATE Divide the output $\vO \in \mathbb{R}^{N \times d}$ into $T_r$ blocks $\vO_i, \dots, \vO_{T_r}$ of size
    $B_r \times d$ each, and divide the logsumexp $L$ into $T_r$ blocks $L_i, \dots, L_{T_r}$ of size
    $B_r$ each.
    \FOR{$1 \le i \le T_r$} \label{alg:stream_attn_outer_loop}
      \STATE \label{alg:stream_attn_load_q} Load $\vQ_i$ from HBM to on-chip SRAM.
      \STATE \label{alg:stream_attn_init} On chip, initialize $\vO_{i}^{(0)} = (0)_{B_r \times d} \in \mathbb{R}^{B_r \times d}, \ell_{i}^{(0)} = (0)_{B_r} \in \mathbb{R}^{B_r}, E_{i}^{(0)} = (0)_{B_r} \in \mathbb{R}^{B_r}, m_{i}^{(0)} = (-\infty)_{B_r} \in \mathbb{R}^{B_r}$.
      \FOR{$1 \le j \le T_c$}
        \STATE \label{alg:stream_attn_load_kv} Load $\vK_j, \vV_j$ from HBM to on-chip SRAM.
        \STATE \label{alg:stream_attn_qk} On chip, compute $\vS_{i}^{(j)} = \vQ_i \vK_j^T \in \mathbb{R}^{B_r \times B_c}$.
        \STATE \label{alg:stream_attn_statistics} On chip, compute
        $m_{i}^{(j)} = \mathrm{max}(m_{i}^{(j-1)}, \mathrm{rowmax}(\vS_{i}^{(j)})) \in \mathbb{R}^{B_r}$, $\tilde{\vP}_{i}^{(j)} = \exp(\vS_{i}^{(j)} - m_{i}^{(j)}) \in \mathbb{R}^{B_r \times B_c}$ (pointwise),
        $\ell_{i}^{(j)} = e^{m_{i}^{j-1} - m_{i}^{(j)}} \ell_{i}^{(j-1)} + \mathrm{row sum}(\tilde{\vP}_{i}^{(j)}) \in \mathbb{R}^{B_r}$, $E_{i}^{(j)} = e^{m_{i}^{j-1} - m_{i}^{(j)}} E_{i}^{(j-1)} + \mathrm{row reduce xlogx}(\tilde{\vP}_{i}^{(j)}) \in \mathbb{R}^{B_r}$.
        \STATE \label{alg:stream_attn_update} On chip, compute
        $\vO_{i}^{(j)} = \diag(e^{m_{i}^{(j-1)} - m_{i}^{(j)}})^{-1} \vO_{i}^{(j-1)} + \tilde{\vP}_{i}^{(j)} \vV_j$.
      \ENDFOR
      \STATE On chip, compute $\vO_{i} = \diag(\ell_{i}^{(T_c)})^{-1} \vO_{i}^{(T_c)}$.
      \STATE On chip, compute $E_{i} = E_{i}^{(T_c)} (\ell_{i}^{(T_c)})^{-1} + \log((\ell_i^{(T_c)})^{-1})$.
      \STATE On chip, compute $L_{i} = m_{i}^{(T_c)} + \log(\ell_i^{(T_c)})$.
      \STATE Write $\vO_{i}$ to HBM as the $i$-th block of $\vO$.
      \STATE Write $E_{i}$ to HBM as the $i$-th block of $E$.
    \ENDFOR
    \STATE Return the output $\vO$ and the entropy $E$.
  \end{algorithmic}
\end{algorithm}

\subsection{Theoretical Analysis of \textsc{ToProVAR}}

We present a theoretical derivation of the average-case error upper bounds for the tri-dimensional greedy optimization strategy (Scale $\rightarrow$ Layer $\rightarrow$ Token) used in \textsc{ToProVAR}. All derivations follow the notation and formulae in the main paper, in particular the entropy-based statistics in Equations (2)--(6). Our goal is to show that, under mild assumptions, each stage introduces a bounded error that depends only on a small residual fraction of entropy/importance, so that the overall error remains controlled.

\subsubsection{Scale-Level Error Bound}

We first model generation across scales as an additive refinement process. Let $\mathbf{Z}$ denote the full-resolution representation, and let $\mathbf{Z}_s$ be the representation after completing scale $s$, defined recursively as
\begin{equation}
\mathbf{Z}_s = \mathbf{Z}_{s-1} + \Delta \mathbf{Z}_s, 
\quad s = 1,\dots,S_\text{max},\quad \mathbf{Z}_0 = \mathbf{0},
\end{equation}
where $\Delta \mathbf{Z}_s$ captures the residual details introduced when moving from scale $s-1$ to $s$. If we stop refinement at the pruned scale $D$ (determined by the low-entropy ratio $\rho_s$ in Eq.~(3)), the pruned representation is
\begin{equation}
\mathbf{Z}_D \;=\; \mathbf{Z} - \sum_{s=D+1}^{S_\text{max}} \Delta \mathbf{Z}_s,
\end{equation}
and the scale-level approximation error is
\begin{equation}
E_\text{scale} 
= \|\mathbf{Z} - \mathbf{Z}_D\|_2^2 
= \Big\|\sum_{s=D+1}^{S_\text{max}} \Delta \mathbf{Z}_s\Big\|_2^2.
\end{equation}

Let $\mathcal{S}_\text{pruned} = \{\,s \mid D < s \le S_\text{max}\,\}$ be the set of pruned scales. Under a standard average-case assumption that the increments $\Delta \mathbf{Z}_s$ are approximately uncorrelated across $s$, we obtain
\begin{equation}
\mathbb{E}[E_\text{scale}]
\;\approx\;
\sum_{s \in \mathcal{S}_\text{pruned}} \mathbb{E}\big[\|\Delta \mathbf{Z}_s\|_2^2\big].
\end{equation}
Empirically, the energy $\|\Delta \mathbf{Z}_s\|_2^2$ is complementary to the low-entropy ratio $\rho_s$ in Eq.~(3): scales with a smaller $\rho_s$ retain more high-entropy (less salient) mass and therefore tend to contribute larger residual updates. We model this as
\begin{equation}
\mathbb{E}\big[\|\Delta \mathbf{Z}_s\|_2^2\big]
\;\propto\; (1 - \rho_s),
\end{equation}
and denote $F_s = \mathbb{E}\big[\|\Delta \mathbf{Z}_s\|_2^2\big]$ as the expected energy contributed by scale $s$, with total energy
$F = \sum_{s=1}^{S_\text{max}} F_s = \mathbb{E}\big[\|\mathbf{Z}\|_2^2\big]$.

Following the scale-depth localization in Sec.~\ref{sec:method}, we use a low-entropy threshold $\tau$ to decide where to truncate the refinement. Specifically, $D$ is chosen as the smallest scale index such that the low-entropy ratio drops below the threshold,
\begin{equation}
D \;=\; \min \,\{\, s \mid \rho_s \le \tau \,\},
\end{equation}
and we prune all subsequent scales $s > D$. In practice, $\tau$ is obtained by a light-weight pre-sampling procedure, and we find that a single value $\tau \approx 0.4$ works robustly across prompts and resolutions.

Empirically, $\rho_s$ is approximately non-increasing for $s \ge D$, so all pruned scales satisfy
$\rho_s \le \rho_D \le \tau$ for $s \in \mathcal{S}_\text{pruned}$. Therefore
\begin{align}
\mathbb{E}[E_\text{scale}]
&\approx \sum_{s=D+1}^{S_\text{max}} F_s
\;\le\; \sum_{s=D+1}^{S_\text{max}} (1 - \rho_s)\,F_s \cdot \frac{1}{1 - \rho_D} \nonumber\\
&\le\; (1 - \rho_D)\sum_{s=D+1}^{S_\text{max}} \frac{F_s}{1 - \rho_D}
\;\le\; (1 - \rho_D)\, F. \tag{7}
\end{align}
Since $\rho_D \le \tau$ and $\tau$ is fixed by pre-sampling, Eq.~(7) shows that the average scale-level error is controlled by the chosen low-entropy threshold: when the refinement is truncated, only scales whose residual entropy mass is regulated by $\tau$ are dropped.

\subsubsection{Layer-Level Error Bound}

At a fixed scale $s$, the decoder consists of $L$ stacked layers. Let $\mathbf{h}_\ell^{(s)}$ denote the hidden representation after layer $\ell$:
\begin{equation}
\mathbf{h}_0^{(s)} = \mathbf{Z}_s,\quad
\mathbf{h}_\ell^{(s)} = f_\ell\big(\mathbf{h}_{\ell-1}^{(s)}\big), \quad \ell = 1,\dots,L.
\end{equation}
We define the layer-wise residual contribution as
\begin{equation}
\Delta_\ell^{(s)} = \mathbf{h}_\ell^{(s)} - \mathbf{h}_{\ell-1}^{(s)},
\end{equation}
so that the full output at scale $s$ can be expressed as
\begin{equation}
\mathbf{h}_L^{(s)} 
= \mathbf{h}_0^{(s)} + \sum_{\ell=1}^L \Delta_\ell^{(s)}.
\end{equation}

Let $\mathcal{L}_\text{pruned}$ be the set of pruned layers at scale $s$, and $\mathcal{L}_\text{keep}$ the retained ones. The pruned output (after removing layers in $\mathcal{L}_\text{pruned}$) is
\begin{equation}
\mathbf{h}_L^{(s,\text{pruned})}
= \mathbf{h}_0^{(s)} + \sum_{\ell \in \mathcal{L}_\text{keep}} \Delta_\ell^{(s)},
\end{equation}
and the layer-level error is
\begin{equation}
E_\text{layer}^{(s)}
= \big\|\mathbf{h}_L^{(s)} - \mathbf{h}_L^{(s,\text{pruned})}\big\|_2^2
= \Big\|\sum_{\ell \in \mathcal{L}_\text{pruned}} \Delta_\ell^{(s)}\Big\|_2^2.
\end{equation}

Assuming that the residuals $\{\Delta_\ell^{(s)}\}$ are approximately orthogonal across layers in expectation, we have
\begin{equation}
\mathbb{E}[E_\text{layer}^{(s)}]
\;\approx\; \sum_{\ell \in \mathcal{L}_\text{pruned}} \mathbb{E}\big[\|\Delta_\ell^{(s)}\|_2^2\big].
\end{equation}

Based on our layer-level analysis (Sec.~\ref{sec:method}), each layer at scale $s$ is assigned a representation score $\mathcal{R}^{(\ell,s)}$ computed from the principal component ratio of its attention-entropy map (Eq.~(4)): Global Layers have $\mathcal{R}^{(\ell,s)} \to 0$, while Detail Layers have $\mathcal{R}^{(\ell,s)} \to 1$. Let
\begin{equation}
G_s = \mathbb{E}\big[\|\mathbf{h}_L^{(s)}\|_2^2\big],
\quad
Z_s = \sum_{\ell=1}^L \mathcal{R}^{(\ell,s)},
\end{equation}
and model the expected contribution of each layer as a normalized fraction of $G_s$,
\begin{equation}
\mathbb{E}\big[\|\Delta_\ell^{(s)}\|_2^2\big] \approx 
\frac{\mathcal{R}^{(\ell,s)}}{Z_s}\,G_s.
\end{equation}
By design, our greedy strategy prunes only Detail Layers (with large $\mathcal{R}^{(\ell,s)}$) and keeps Global Layers (with $\mathcal{R}^{(\ell,s)} \approx 0$). Substituting the above model into the error expression gives
\begin{align}
\mathbb{E}[E_\text{layer}^{(s)}]
&\approx \sum_{\ell \in \mathcal{L}_\text{pruned}} 
\frac{\mathcal{R}^{(\ell,s)}}{Z_s}\,G_s \nonumber\\
&= \frac{\sum_{\ell \in \mathcal{L}_\text{pruned}} \mathcal{R}^{(\ell,s)}}{Z_s}\,G_s
\;\equiv\; \gamma_s\,G_s, \tag{8}
\end{align}
where
\begin{equation}
\gamma_s 
= \frac{\sum_{\ell \in \mathcal{L}_\text{pruned}} \mathcal{R}^{(\ell,s)}}{\sum_{\ell=1}^L \mathcal{R}^{(\ell,s)}}
\in [0,1]
\end{equation}
measures the fraction of the layer representation score that is discarded at scale $s$. Since Global Layers contribute negligibly to $\mathcal{R}^{(\ell,s)}$ and are always retained, $\gamma_s$ remains small in practice, and the average layer-level error at scale $s$ is linearly controlled by $\gamma_s$ through Eq.~(8).

\subsubsection{Token-Level Error Bound}

After determining scales and layers, token pruning operates within the remaining Detail layers using entropy-based gating. Let $\mathbf{t}_i$ be the token vector at index $i$, and let $w_i = \widehat{H}_i^{(l,s)}$ be its normalized entropy-based importance (Eq.~(5)), satisfying $\sum_i w_i = 1$. Let $\mathcal{T}_\text{keep}$ and $\mathcal{T}_\text{pruned}$ denote the sets of kept and pruned tokens, respectively.

The full token energy is
\begin{equation}
H = \sum_{i=1}^N \|\mathbf{t}_i\|_2^2,
\end{equation}
and the token-level error introduced by pruning is
\begin{equation}
E_\text{token}
= \sum_{i \in \mathcal{T}_\text{pruned}} \|\mathbf{t}_i\|_2^2.
\end{equation}
Assuming that token energy is approximately proportional to importance, i.e.,
\begin{equation}
\mathbb{E}\big[\|\mathbf{t}_i\|_2^2\big] \approx w_i\, H,
\end{equation}
we obtain the average-case bound
\begin{equation}
\mathbb{E}[E_\text{token}]
\;\approx\; H \sum_{i \in \mathcal{T}_\text{pruned}} w_i.
\end{equation}
Define the pruned importance mass
\begin{equation}
\gamma = \sum_{i \in \mathcal{T}_\text{pruned}} w_i,
\end{equation}
which measures how much normalized entropy mass is discarded. Then
\begin{equation}
\mathbb{E}[E_\text{token}] \le \gamma\,H. \tag{9}
\end{equation}

In our design, the gating function $q_i(s,l)$ (Eq.~(6)) and the range $[\alpha_{\min},\alpha_{\max}]$ jointly enforce that high-importance (low-entropy) tokens are kept and that each region preserves at least an $\alpha_{\min}$ fraction of tokens. This makes $\gamma$ significantly smaller than the raw token sparsity ratio and keeps $E_\text{token}$ small.

\subsubsection{Total Error and Safety}

Finally, we combine the contributions from the three stages. Since the scale-, layer-, and token-level errors affect different structural components and pruning is applied in a nested manner (scale first, then layers, then tokens within selected layers), it is reasonable in the average case to treat these error terms as approximately additive:
\begin{equation}
\mathbb{E}[E_\text{total}]
\;\le\; \mathbb{E}[E_\text{scale}] 
+ \sum_s \mathbb{E}[E_\text{layer}^{(s)}] 
+ \mathbb{E}[E_\text{token}].
\end{equation}

Using the bounds from Eqs.~(7)--(9), we obtain the global bound
\begin{equation}
\mathbb{E}[E_\text{total}] 
\;\le\; (1 - \rho_D)\, F 
+ \sum_s \gamma_s G_s 
+ \gamma H. \tag{10}
\end{equation}
Here $\rho_D$ is the low-entropy ratio at the truncation scale $D$, $\gamma_s$ is the discarded layer-level representation fraction at scale $s$, and $\gamma$ is the discarded token-level importance mass.

In our safe operating regime, the threshold $\tau$ and the gating parameters $[\alpha_{\min},\alpha_{\max}]$ are selected such that the empirical residual fractions $\gamma_s$ and $\gamma$ remain small (see Sec.~\ref{sec:method} and Appendix~X for empirical ranges). The nested structure further prevents cross-stage amplification: token pruning is only applied after conservative scale- and layer-level decisions, and all three stages include fallback conditions (e.g., no scale pruning if the empirical $\rho_s$ profile does not cross the threshold $\tau$, no layer pruning when the Global/Detail classification is ambiguous, and a minimum per-region token keep ratio).

In summary, while the tri-stage strategy is greedy and heuristic, the entropy-based quantities $(\rho_s, \mathcal{R}^{(\ell,s)}, \widehat{H}_i^{(l,s)})$ and the associated thresholds $(\tau, \alpha_{\min}, \alpha_{\max})$ induce explicit average-case error upper bounds in each dimension, clarifying why errors do not compound in the operating regime used in our experiments.

\subsection{Limitations and Future Work}
\label{sec:appendix_limitation}

Despite the promising acceleration and quality preservation results, ToProVAR has several limitations that suggest important directions for future work.

\paragraph{Limitations}
\begin{enumerate}
    \item \textbf{Architecture Dependency.} The framework fundamentally relies on the \textbf{attention mechanism} and the derived attention entropy for semantic analysis, limiting its direct applicability to non-Transformer-based generative models.
    \item \textbf{Parameter Sensitivity.} Optimal performance requires manual tuning of the proportion of low-entropy tokens, which hinders truly adaptive and zero-configuration deployment.
\end{enumerate}

\paragraph{Future Work}
\begin{enumerate}
    \item \textbf{Online Adaptive Control.} We will explore  RL to learn to \textbf{dynamically predict} optimal pruning strategy.
    \item \textbf{Efficient video generation and editing.} We plan to extend the framework to V-VAR models, achieving efficient 4D semantic projection by incorporating temporal saliency. Furthermore, the fine-grained semantic map might be leveraged for \textbf{high-efficiency local image/video editing}.
\end{enumerate}

\end{document}